\documentclass{article}
\usepackage[nonatbib,final]{neurips_2021}

\usepackage{amssymb}
\usepackage{amsfonts}

\usepackage[bb=px]{mathalpha}
\usepackage{graphicx}
\usepackage[caption=false]{subfig}
\usepackage{bm}
\usepackage{newfloat}
\usepackage{algorithm}
\usepackage{algorithmic}
\usepackage{booktabs}
\usepackage{dsfont}
\usepackage{blindtext} 
\usepackage{array,booktabs}
\newcolumntype{L}{@{}>{\kern\tabcolsep}l<{\kern\tabcolsep}}
\usepackage{colortbl}
\usepackage{xcolor}
\usepackage{hyperref}
\usepackage{soul}

\usepackage{amsthm}
\usepackage{amsmath}

\begin{document}

\title{AA-Forecast: Anomaly-Aware Forecast for Extreme Events}
\author{
  Ashkan Farhangi \\
  University of Central Florida\\
  Orlando, USA \\
  \texttt{ashkan.farhangi@ucf.edu} \\
  \And
  Jiang Bian \\
  Baidu Inc.\\
  Beijing, China \\
  \texttt{bianjiang03@baidu.com} \\
  \AND
  Arthur Huang \\
  University of Central Florida \\
  Orlando, USA \\
  \texttt{arthur.huang@ucf.edu} \\
  \And
  Haoyi Xiong \\
  Baidu Inc. \\
  Beijing, China \\
  \texttt{xionghaoyi@baidu.com} \\
   \And
   Jun Wang \\
   University of Central Florida \\
   Orlando, USA \\
   \texttt{jun.wang@ucf.edu} \\
   \And
   Zhishan Guo \\
    University of Central Florida\\
    Orlando, USA \\
   \texttt{zhishan.guo@ucf.edu} \\   
}



\maketitle

\begin{abstract}
Time series models often deal with extreme events and anomalies, both prevalent in real-world datasets. Such models often need to provide careful probabilistic forecasting, which is vital in risk management for extreme events such as hurricanes and pandemics.
However, it is challenging to automatically detect and learn to use extreme events and anomalies for large-scale datasets, which often require manual effort.
Hence, we propose an anomaly-aware forecast framework that leverages the previously seen effects of anomalies to improve its prediction accuracy during and after the presence of extreme events.
Specifically, the framework automatically extracts anomalies and incorporates them through an attention mechanism to increase its accuracy for future extreme events.
Moreover, the framework employs a dynamic uncertainty optimization algorithm that reduces the uncertainty of forecasts in an online manner.
The proposed framework demonstrated consistent superior accuracy with less uncertainty on three datasets with different varieties of anomalies over the current prediction models.
\end{abstract}

\section{Introduction}
Climate change is increasing the severity of natural disasters. Compared to the 1990s, natural disasters quadrupled in terms of economic damage in the U.S. alone~\cite{noaa2021}.
Time series forecasting during the presence of such extreme events (e.g., hurricanes) is critical for resource allocation and resilience planning~\cite{ecmlts4,resourceallocationexample,resilienceexample}. 

Intuitively, the high accuracy and low uncertainty of the forecasts are critical insights for uncovering the influence of external shocks and events on large-scale time series data~\cite{ecmlts1}.
To provide sustainable economic development and resilience planning, it is crucial to understand how different industries are influenced by and recover from such extreme events over time~\cite{andrewClimate}. 
However, it remains a challenge to develop reliable and accurate prediction models as the real-world dataset often contains anomalies that tend to be rare and random. Hence, it is crucial to develop a forecast model that can leverage previously seen extreme events and anomalies for future forecasts. 

Although there have been considerable achievements in machine learning-based models, existing methods tend to overlook anomalies' special effects on real-world time series data. For instance,
LSTMs~\cite{lstmog} address the vanishing gradient problem via gate mechanism and have the ability to capture complex temporal dependencies~\cite{uber}. Yet, Khandelwal et al.~\cite{lstmnotlongdepend} show that even LSTMs have a limited ability to capture long-term dependencies, and their awareness of context degrades as
the length of the input sequence increases. Consequently, making them inefficient to capture and learn from rare occurrences or extreme events.

As an alternative, Li et al.~\cite{nips_ts_attention_locaclity} considered the use of transformers for time series prediction. Transformers use a self-attention mechanism that allows each observation in the feature sequence to attend independently to every other feature in the sequence. However, they have considerable computational and memory requirements that grow quadratically with respect to sequence length, making it computationally rigorous to train large-scale data~\cite{nips_ts_attention_locaclity}. Such deficiency makes them computationally unsuitable for extreme events that often appear in longer sequences than the transformer's inputs. Moreover, it was not even clear from the design itself that transformers can be as effective as RNNs, whereas Zaheer et al.~\cite{bigbird} reported that the attention mechanism in transformers does not even obey the sequence order of time steps which is essential for the time series domain. What is more, non-transformer architectures (i.e., MLP) when designed and trained properly, can perform competitively with transformers~\cite{transformerdownfallMLP}.

This lack of systematic strategy in handling anomalies and not providing predictions with nontransparent uncertainty levels makes current forecasting methods unreliable during the presence of extreme events.
As a result, a key aspect of our knowledge in developing time series models for critical moments of extreme events will remain a puzzle unless the long-term effects of anomalies are well captured and utilized. 

\vspace{2mm}
\noindent {\bf Contribution.} This work proposes a novel and generalized anomaly-aware prediction framework, AA-Forecast, which automatically extracts and uses anomalies to optimize its probabilistic forecasting. Specifically, 

\begin{itemize}
 \item[$\bullet$] AA-Forecast extracts anomalies through a novel decomposition method and leverages them through an attention mechanism designed to optimize its probabilistic forecasting during extreme events. Also, AA-Forecast is able to perform zero-shot prediction for unseen time series and does not suffer from quadratic computational time and memory complexity of transformers.

 \item[$\bullet$] An online optimization procedure is proposed to minimize the prediction uncertainties of AA-Forecast framework, which features applying the optimal dropout probability at every time step during testing.

 \item[$\bullet$] Extensive experimental studies are conducted on three real-world datasets prone to extreme events and anomalies. The comparisons with state-of-the-art models illustrate the higher accuracy and less uncertainty in the AA-Forecast's prediction. 
\end{itemize}

\section{Problem Formulation}
\label{section-preliminary}

In this study, we are interested in the task of time series forecast under the influence of extreme events and anomalies. 
Mathematically, given a dataset $\mathbf{D} = \{\mathbf{x}^{(1)},\mathbf{x}^{(2)},\ldots,\mathbf{x}^{(K)}\}$ with $K$ univariate time series, $\mathbf{x}^{(k)}=\{x^{(k)}_{1}, x^{(k)}_{2},\ldots, x^{(k)}_{T}\}$ denotes a time series instance with length $T$, where $\mathbf{x}^{(k)} \in \mathbb{R}^{T}$. For every time step, the corresponding extreme events are aligned and labeled as covariates $\mathbf{e}^{(k)} = \{e^{(k)}_1, e^{(k)}_2,\ldots, e^{(k)}_{T}\}$. Extreme events are considered as the influence of external events that promote a dynamic occurrence within a limited time steps~\cite{extremeeventdefinition}. Specifically, $e^{(k)}_{t} \in \mathbb{R}$ indicates the level of extreme event (e.g., hurricane category) at time $t$, otherwise, $e^{(k)}_{t} = 0$ indicates a non-extreme event condition for periods outside of the event. 
To this end, we denote the data with extreme events as a series of tuples $\widehat{\mathbf{x}}^{(k)} \overset{\Delta}{=} \{(x^{(k)}_1,e^{(k)}_1),(x^{(k)}_2,e^{(k)}_2),\ldots, (x^{(k)}_{T},e^{(k)}_{T})\}$. 
Particularly, given the previous $\tau$ observations $\widehat{\mathbf{x}}_{t-\tau+1:t}^{(k)} = \{({x}^{(k)}_{t-\tau+1},e^{(k)}_{t-\tau+1}), ({x}^{(k)}_{t-\tau+2},e^{(k)}_{t-\tau+2}),\ldots, ({x}^{(k)}_{t},e^{(k)}_{t})\}$, we aim to model the conditional distribution of the next observation:

\begin{equation}
{p}(x^{(k)}_{t+1}\ | \widehat{\mathbf{x}}^{(k)}_{t-\tau+1:t} ; \mathbf{\Phi}),
\end{equation}
where $\mathbf{\Phi}$ denotes the parameters of a nonlinear prediction model. We are also interested in reducing the uncertainty of predictions in an online setting, whereas uncertainty of prediction can be viewed as the variability of the distribution. Therefore, the optimization problem during the online settings is defined as:
\begin{equation}
 \mathbf{\Phi}^{*}_{\text{on}} = \text{argmin}_{\mathbf{\Phi}} \: \mathcal{V} \left({p}(x^{(k)}_{t+1}\ | \ \widehat{\mathbf{x}}^{(k)}_{t-\tau+1:t} ; \mathbf{\Phi})\right),
\end{equation}
where $\mathcal{V}\left(\cdot\right)$ represents the variability of the probability distribution and $\mathbf{\Phi}^{*}_{\text{on}}$ is the optimal online parameters of the nonlinear prediction model that produces the least amount of uncertainty in each time step.

\section{AA-Forecast Framework}
\label{section-methodology}
The proposed AA-Forecast framework consists of three main components. 
Section~\ref{sec-AnomalyDecomposition} proposes a novel anomaly decomposition method that automatically extracts the anomalies and essential features of the time series data. Then, the extracted anomalies are fed into an anomaly-aware model detailed in Section~\ref{sec-Anomaly-Aware}. Specifically, it leverages an attention mechanism on anomalies and extreme events to produce the distribution of the forecasts. To further reduce the forecast uncertainty in an online manner, Section~\ref{sec-Dynamic} proposes a dynamic uncertainty optimization algorithm.

\subsection{STAR Decomposition}
\label{sec-AnomalyDecomposition}

STAR decomposition is used as a strategy to not only extract the anomalies and sudden changes of data but also decompose the complex time series to its essential components. 
Unfortunately, widely popular decomposition method such as STL~\cite{stlog} does not extract anomalies. Although recent works such as STR~\cite{strdecomposition21} and RobustSTL~\cite{wen2019robuststl} are designed to be robust to the extreme effect of anomalies in their decomposition, they are not used to explicitly extract anomalies from the residual component.

To alleviate these issues, we propose STAR decomposition that are decomposes the original time series $\mathbf{x}^{(k)}$ in a multiplicative manner to its \textbf{s}easonal ($\mathbf{s}^{(k)}$), \textbf{t}rend ($\mathbf{t}^{(k)}$), \textbf{a}nomalies ($\mathbf{a}^{(k)}$), and \textbf{r}esidual ($\mathbf{r}^{(k)}$) components: 
\begin{equation}
\mathbf{x}^{(k)} =
\mathbf{s}^{(k)} \times
\mathbf{t}^{(k)} \times
\mathbf{a}^{(k)} \times 
\mathbf{r}^{(k)}
\end{equation}

\begin{figure}[t]
 \centering
 \includegraphics[trim={1.2cm 3cm 1.2cm 4.8cm},clip,width =1\linewidth]{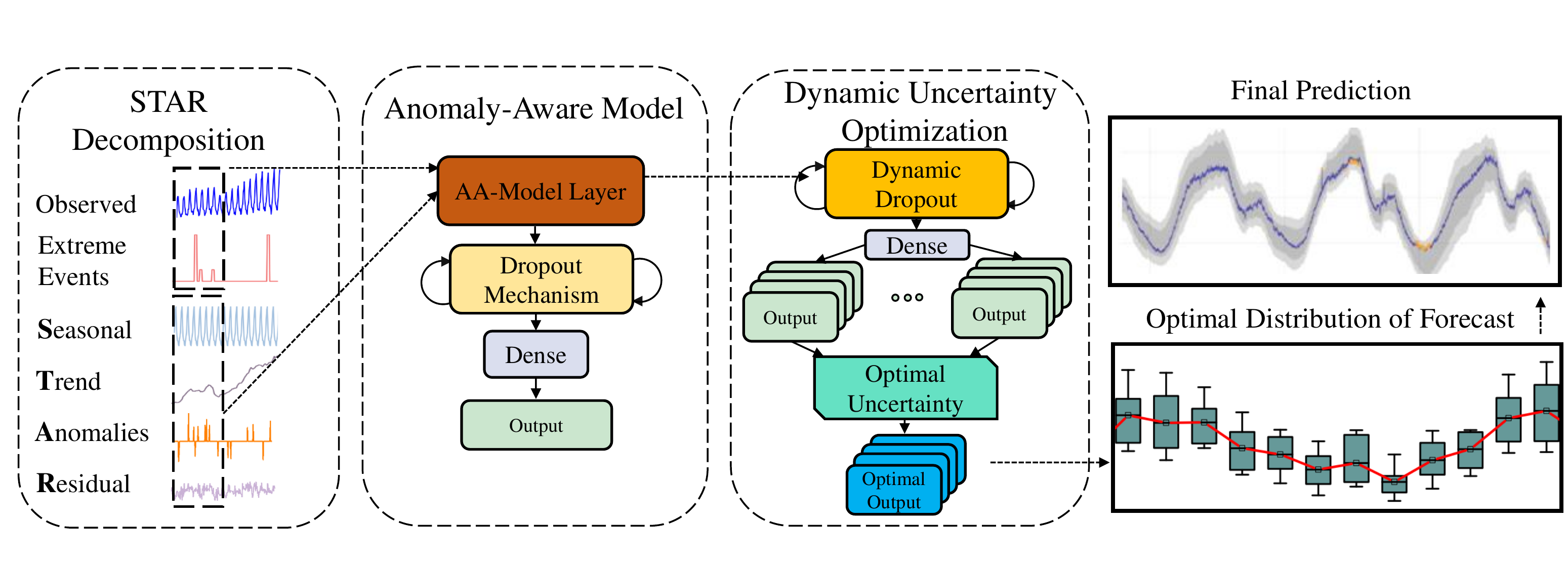}
 \caption{Main components of AA-Forecast: (i) \textbf{STAR Decomposition} to automatically extract essential features such as anomalies, (ii) an \textbf{Anomaly-Aware Model} to leverage such extracted features, and (iii) a \textbf{Dynamic Uncertainty Optimization} to reduce the uncertainty of the network. The final predicted series contains confidence intervals with the least uncertainty.}
 \label{fig_framework}
\end{figure}

Such decomposition not only increases the dimensions of the original data but also supplies us with the extracted anomalies.
As shown in Figure~\ref{fig_framework}, we begin the decomposition by approximating the trend line $\mathbf{t}^{(k)}$ with the locally weighted scatterplot smoothing (i.e., LOESS ~\cite{loessog}).
Then, we divide the original data $\mathbf{x}^{(k)}$ by the approximated trend line to derive the detrended time series\footnote{We use the log transform of $\mathbf{x}^{(k)}$ to handle the situation that specific values of original data are zero.}. 
We then partition the detrended time series into periods of cyclic sub-series where the cycle size is determined by the time interval of the dataset. As an example, the cycle size for a monthly dataset would be 12 (one year as a cycle). Then, we obtain the seasonal component ($\mathbf{s}^{(k)}$) by grouping the detrended series in each period and deriving the average value of each period across the time series. Subsequently, the residual component ($\mathbf{r}^{(k)}$) is derived by dividing the seasonal and trend segments from the original series. 

Note that the anomaly component ($\mathbf{a}^{(k)}$) can be considered as the oddities of the dataset, which do not follow the extracted trend or seasonal components. Intuitively, anomalies are spread out through residual components, which also contain noise and other real-world effects. To distinguish the anomalies from residual components, statistical metrics such as mean and variance are not the appropriate measure as they are highly sensitive to the severity level of anomaly values. As one expects, the severity of the anomalies can change the mean and variance values which are unwanted. To resolve this issue, we leverage the median of the residuals, which is immune to the severity of the outliers in the residual components. Next, we define robustness score $\rho^{(k)}_t$ for each observation at time $t$ as: 
\begin{equation}
\rho_t^{(k)} = \frac{|r_t^{(k)} - \Dot{r}^{(k)}|}{\sqrt{\frac{\sum_{t=1}^{T}|r^{(k)}_t-\Dot{r}^{(k)}|}{T-1}}} 
\end{equation}
where $\rho_t^{(k)}$ stands for the strength of the anomalies, $r_t^{(k)}$ is the residual at time step $t$ and $\Dot{r}^{(k)}$ is the median of the residuals.
Note that the larger $\rho_t$ indicates that a drastic change has occurred in the trend and seasonal components.
We then extract the anomalies from residuals as below:
\begin{equation}
\mathbf{a}_t^{(k)}=\left\{\begin{array}{ll}
1, & {\rho}_t^{(k)} < \rho_{c}^{(k)} \\
r_t^{(k)}, & \rho_t^{(k)} > \rho_{c}^{(k)}
\end{array}\right.
\end{equation}
where $\rho_{c}^{(k)}$ is the constant threshold given by the value of a robustness score ranked in the $p$-value 0.05\footnote{Adopted based on the choice of the $p$-value (0.05) which is used as a standard level of statistical significance.} while the values of elements in $\rho^{(k)}$ are ranked in descending order from large to small. 
Notably, when the value of the anomaly component ($\mathbf{a}^{(k)}$) deviates further from the value $1$, it indicates an abrupt change in the trend and the seasonal component (no sign of anomalies). On the contrary, when both anomaly and residual values are equal 1 ($\mathbf{r}_{t}^{(k)}=1$ and $\mathbf{a}_{t}^{(k)}=1$), it indicates that the observed signal at time $t$ explicitly follows the trend and seasonal component. Note that such important information might not be automatically inferred when additive decomposition methods are being used. This is due to the fact that the values of residual components can differ from one dataset to another which requires manual effort in their detection.
 
A sample result of anomaly decomposition is shown in the left-most part of Figure~\ref{fig_framework}, where the observed time series data is decomposed into its seasonal, trend, anomalies, and residual components respectfully. Each of these components holds essential information about the characteristics of the time series and will be leveraged to train the forecast model. To this end, we concatenate the derived decomposed vector of time series with the input, which includes the observed time series and its labeled extreme event. Specifically, STAR decomposition concatenates the original time series to $\widetilde{\mathbf{x}}^{{(k)}} = (\mathbf{x}^{(k)},\mathbf{e}^{(k)},\mathbf{s}^{(k)},\mathbf{t}^{(k)},\mathbf{a}^{(k)},\mathbf{r}^{(k)})$ which can be leveraged by the anomaly-aware model described in the next section.

\subsection{Anomaly-Aware Model}
\label{sec-Anomaly-Aware}
The Anomaly-Aware model is designed to explicitly incorporate extracted anomalies $\mathbf{a}^{(k)}$ and extreme event covariates $\mathbf{e}^{(k)}$ into the prediction. 
As these features are rare in the whole time series, feeding them directly into a regular RNN like LSTM~\cite{lstmog} can be potentially ignored during the training of the model. Note that the extracted anomalies and previously experienced external events hold valuable information regarding the effect of extreme events that should be handled carefully.

Recent robust prediction models rely on the LSTMs or transformers architecture to provide robustness in their prediction. Even though LSTMs are designed to obtain long-term dependencies, their capacity to pay different degrees of attention to sub-window features within large time steps is inadequate~\cite{bigbird}. As an example, Khandelwa et al.~\cite{lstmnotlongdepend} showed that even though the LSTM model can have an effective sequence size of $200$ observations, they are only able to sharply distinguish the $50$ closest observations. This indicates that even LSTMs struggle to capture long-term dependencies.
On the other hand, conventional transformers 
suffer from quadratic computation and memory requirements, which limits their ability to process long input sequences.
Even though such memory bottlenecks have been improved by using sparse-attention algorithms~\cite{nips_ts_attention_locaclity}, their performance improvement is not significant compared to a full-attention mechanism for real-world datasets~\cite{lim2019temporal}. Given that extreme events and anomalies are rare and can appear at very long distances from each other, it is computationally infeasible to increase the input sequence to provide attention to all previously seen anomalies and extreme events. 

To address such problems, one must pay attention to all of the anomalies and extreme events throughout the dataset, no matter how far they have occurred. Intuitively, due to their rare nature, they hold greater importance in learning, given that the trend and seasonal patterns are often easier to predict by statistical or deep learning models. Ergo, we developed a novel attention mechanism explicitly for extreme events and anomalies, which are considered the crucial time steps of time series data and often cause the biggest error in prediction.

\noindent\textbf{Architecture Design of AA-Model.}
LSTMs and GRUs are suitable for predicting the recurring patterns with a fairly low computational time and memory complexity which suffer from the quadratic complexity of full-attention transformers. 
However, we enhance the long-term dependencies of these models through an attention mechanism that retains the effect of anomalies and extreme events for future predictions. Such a decision in architecture allows the model not only to be computationally feasible for handling large-scale datasets but also to take the critical moments of extreme events and anomalies into consideration.

\begin{figure}[!t]%
 \centering
 {{\includegraphics[trim={11cm 4.6cm 11.2cm 3.5cm},clip,width =.43\linewidth]{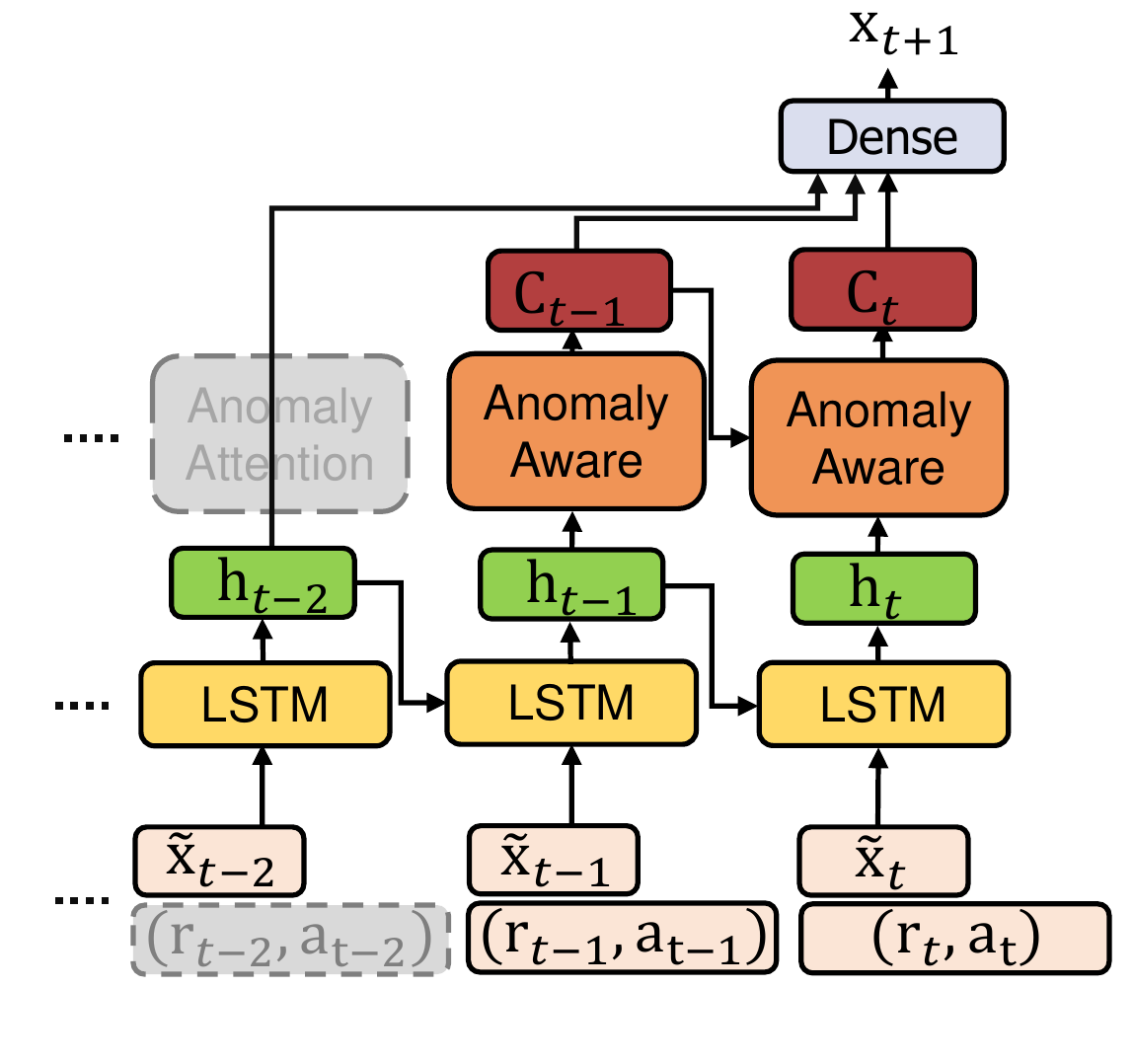} }}%
 \qquad
 {{\includegraphics[trim={7.4cm 4.2cm 8.5cm 3.1cm},clip,width =.51\linewidth]{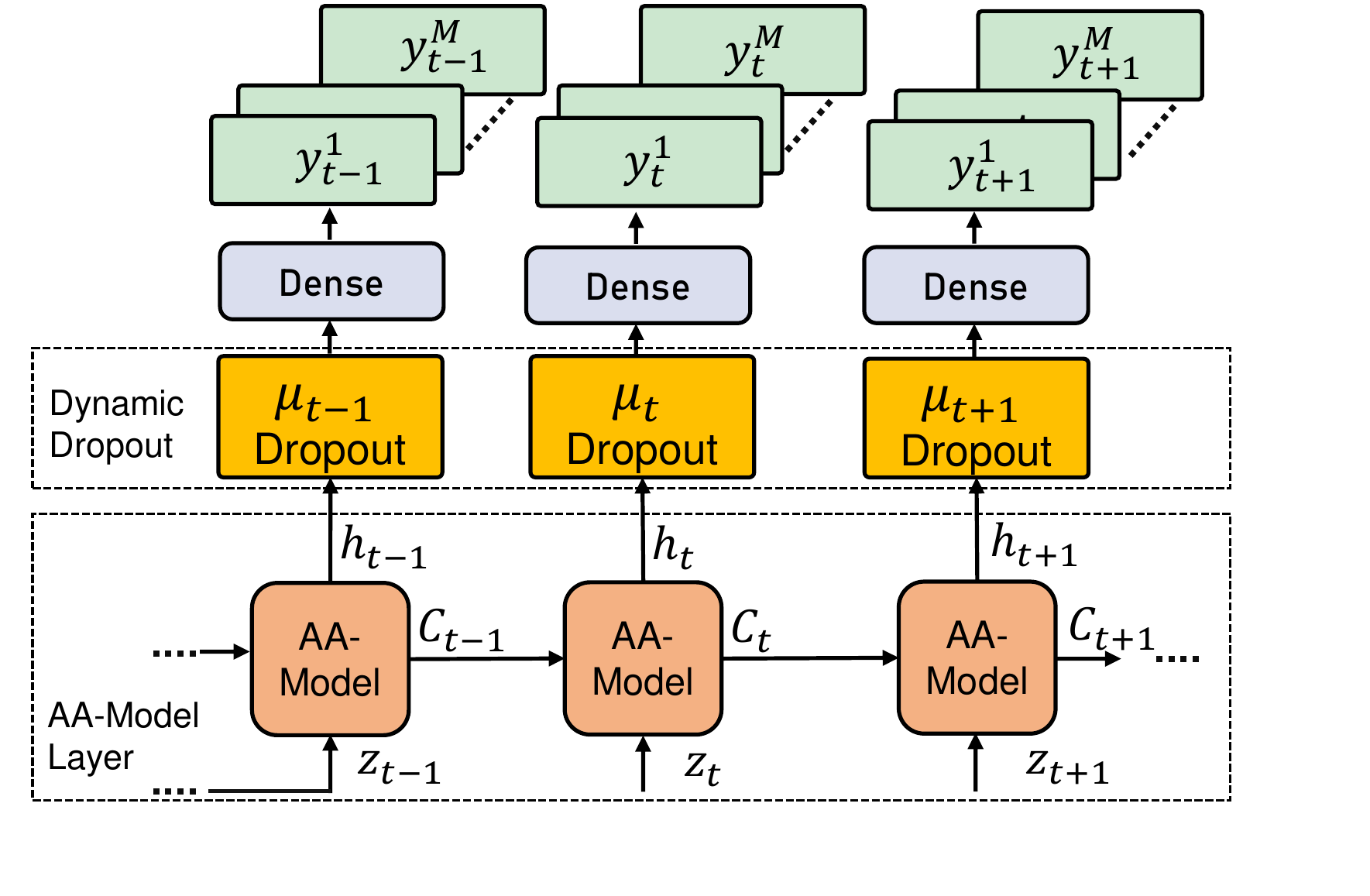} }}%
 \caption{\textbf{Left:} AA-Model Architectures. \textbf{Right:} Dynamic dropout $\mu_{t}$ determines the optimal probability of dropout at each time step during the online settings (i.e., inference). The output $\hat{y}$ consists of a distribution of predicted test values. The dropout optimization improves the certainty and accuracy at each time step $t$ by determining how relevant the previous hidden state is for the next time step prediction.}%
 \label{fig_arch}%
\end{figure}

Given the past $\tau$ time steps of observations as $\Tilde{\mathbf{x}}_{{t:t-\tau+1}}$\footnote{To reduce the ambiguity of the AA-Forecast layer, we are omitting the superscript (k) from this section}, we derive the hidden states of an RNN that deals with vanishing gradient problem (e.g., LSTM or GRU) as:

\begin{equation}
\mathbf{h}_{{t:t-\tau+1}} =\text{RNN}\left(\Tilde{\mathbf{x}}_{{t:t-\tau+1}}\right), \\
\end{equation}
where $\mathrm{h}_{t}$ is the hidden layer of $\text{RNN}$ at time step $t$. Note that we are only giving attention to anomalies and extreme events which are naturally rare and belong to a small population of observations. Moreover, both could have different impacts on the prediction and based on the type of dataset, can be challenging to model. Hence, we design the attention mechanism to automatically incorporate extreme events and anomalies during their occurrence: 
\begin{equation}
 J = \{ t \in \mathbb{Z}^{+} | e_{t} \neq 0 \vee a_{t}\neq 1\},
\end{equation}
where $J$ is the set of time steps including two possible circumstances: presence of extreme events covariates ($e_{t} \neq 0$) or anomalies ($a_{t} \neq 1$).
We then gather all the previous hidden states of the RNNs for all critical time steps in $J$ and regularize them by the weights generated from the attention layer as $v_{t}$ which follows:
\begin{equation}
v_{t} =\text{tanh}(\mathbf{w}_{\alpha}^{\top} \mathbf{h}_{t}+b_{\alpha}), \quad 	\forall t \in J
\end{equation} 
where $\mathbf{w}_{\alpha}$ and $b_{\alpha}$ are the attention layer's weight and bias.
Then, we derive the attention weights of all previous values as:
\begin{equation}
{\alpha}_{t}=\operatorname{Softmax}\left(v_{1}, v_{2}, \ldots, v_{t}\right), \quad \forall t \in J
\end{equation}
where $\alpha_{t}$ is the attention weight at the critical time steps.
The generated attention weights are then used in the AA-Forecast layer as: 
\begin{equation}
\mathcal{A}_{t}=\left\{\begin{array}{ll}
\mathbf{h}_{t}, &\forall t \not\in J\\
\sum_{t \in J} \alpha_{t} \cdot \mathbf{h}_{t}, & \forall t \in J
\end{array}\right.
\end{equation}
where the attention values are only calculated in the presence of anomalies and extreme events as shown in Figure~\ref{fig_arch}. 
The value of the next time step is calculated through a dense layer:
\begin{equation}
 y_{t+1} = \mathbf{w}_d (\mathcal{A}_{t:t-\tau+1})+b_d,
\end{equation}
where $\mathbf{w}_d$ and $b_d$ are the weights and biases of the dense layer.
To train the network, we minimize the prediction loss $\mathcal{L}$ which is defined as follows:
\begin{equation}
\mathbf{\Phi}^{*}_{\text{off}} = \text{argmin}_{\mathbf{\Phi}} \: \mathcal{L}\left(\mathcal{F}_\mathbf{\Phi}(\widetilde{\mathbf{x}}),y\right),
\end{equation}
where $\mathcal{F}_\mathbf{\Phi}$ is the anomaly-aware model and $y$ is the training label, which is the ground truth of the next time step prediction. Note that
 $\mathbf{\Phi}^{*}_{\text{off}}$ represents the optimal model parameters after the offline training phase.

\subsection{Dynamic Uncertainty Optimization}
\label{sec-Dynamic}

Although Monte Carlo (MC) dropout~\cite{mcdropOG} probability is treated as a static hyperparameter in previous studies~\cite{salinas2020deepar,uber0}, it plays an important role in the prediction outcome and can be leveraged to reduce the uncertainty of the prediction during the testing phase~\cite{ecmlts3}. Therefore, we rely on an automatic selection mechanism for the optimal dropout in online settings. Such selection is based on the uncertainty of the prediction produced during the testing phase (Figure \ref{fig_arch}).

Note that the model's uncertainty is desired to be the lowest and as stable as possible in real-world settings.
Therefore, it is essential to optimize further the uncertainty of the model prediction both during the offline training and online testing phase.
Specifically, we apply a dropout operation after every AA-Forecast layer with a specific probability ($p$).

AA-Forecast not only reports the prediction distribution but also provides the point prediction (average of the distribution) and the prediction uncertainty (variability of the distribution). 
Specifically, by producing $M$ forecast for every time step in an online manner (test data $\widetilde{\mathbf{x}}^{*}$) from the previously trained model $\mathcal{F}_\mathbf{\Phi}(\widetilde{\mathbf{x}})$, we obtain $M$ outputs $y^*$ as a from the prediction distribution $\left\{{y}_{(1)}^{*}, \ldots, {y}_{(M)}^{*}\right\}$. Then, the average of the distribution is calculated as 
$
 \bar{{y}}^{*}=\frac{1}{M} \sum_{m=1}^{M} {y}^{*}_{(m)}
$.

We represent uncertainty as to the variability of the prediction distribution ---- the standard deviation (SD) of the probability distribution of future observations conditional on the information available at the time of forecasting. We further optimize the uncertainty of the framework by deriving the optimal dropout probability $p$ at each time step. We derive the prediction error for the probability $p$ between 0 and 1 with 0.1 increments. Notably, without such probability (i.e., $p=0$) the model prediction deviates from probabilistic forecasting and does not provide a level of uncertainty in its prediction for each time step. The optimal uncertainty $\mu_{t}$ is then reported when it results in a minimal variance (i.e., SD) of the predicted values, 
thereby reducing the prediction uncertainty to its minimum during the testing phase. To this end, the prediction uncertainty is formulated as:
\begin{equation}
\sigma^{2}\left(\mathcal{F}_\mathbf{\Phi}(\widetilde{\mathbf{x}}^{*})\right)=\sqrt{\frac{1}{M} \sum_{m=1}^{M}\left({y}_{(m)}^{*}-\bar{{y}}^{*}\right)^{2}}.
\end{equation}

 \begin{algorithm}[!ht]
 \caption{Psuedecode for AA-Forecast}
 \label{alg:self-learning}
 \begin{algorithmic}[1]
 \renewcommand{\algorithmicrequire}{\textbf{Input:} data
 $\widetilde{\mathbf{x}}^{{(k)}} = (\mathbf{x}^{(k)},\mathbf{e}^{(k)},\mathbf{s}^{(k)},\mathbf{t}^{(k)},\mathbf{a}^{(k)},\mathbf{r}^{(k)})$}
 \renewcommand{\algorithmicensure}{\textbf{Output:} prediction $\hat{y}$}
 \REQUIRE 
\STATE {\text{Initialize parameters} $\mathbf{\Phi}$}
\FOR{$k=1$ to $K_{\text{train}}$}
 \STATE {\text{Sample ($\Tilde{\mathbf{x}}^{k},y^{k}$) from training data:}}
\FOR{$b=1$ to $B$}
 \STATE \quad \quad {$\mathbf{\Phi}_{e+1}$ $\leftarrow$ $\mathbf{\Phi}_{e}$ - $\xi$$\cdot$$\nabla$ $\mathcal{L}$($\mathcal{F}_{\mathbf{\Phi}}(\Tilde{\mathbf{x}}^{k})$, $y^k$)}
 \STATE {Update the optimal parameters: \\ \quad \quad $\mathbf{\Phi}$ = $\text{argmin}_{\mathbf{\Phi}}\mathcal{L}$($\mathcal{F}_{\mathbf{\Phi}}(\Tilde{\mathbf{x}}^{k})$, $y^k$)}
 \ENDFOR
\ENDFOR
\STATE {Dynamic Uncertainty optimization: $\mathbf{\Phi}^{*} \leftarrow \mathbf{\Phi}$}
\FOR{$\delta = 0.1$ \text{to} $0.9$ increment by $0.1$}
\STATE Update the optimal uncertainty: \\ \quad \quad $\mathbf{\Phi}^{*}$ = $\text{argmin}_{\mathbf{\Phi}}\mathcal{V}$($\mathcal{F}_{\mathbf{\Phi}}(x^k)$)
\ENDFOR
 \end{algorithmic}
 \end{algorithm}

Algorithm~\ref{alg:self-learning} presents the pseudocode for AA-Forecast. Specifically, we sample ($\Tilde{\mathbf{x}}^{k},y^{k}$) as a driving example which includes extracted anomalies $\mathbf{a}^{(k)}$ and extreme events $\mathbf{r}^{(k)}.$ Next, we train the model by maximizing the overall prediction accuracy. Upon testing, the network leverage dynamic uncertainty optimization further optimizes the prediction uncertainty automatically in online testing so that it would not require any further training.

Note that the network's predictions during the testing phase cannot benefit from the supervised training. However, the control of variability is possible and ensures that the prediction uncertainty is minimal in each step of future predictions, regardless of whether the labels are provided or not. 
Additionally, the algorithm testing time complexity is similar to other RNN-based models due to the use of dynamic uncertainty optimization during the test phase solely. This allows the model to provide the least amount of uncertainty during the presence of anomalies or extreme events where critical online decisions are being made. 

\begin{figure}[!ht]
 \centering
 \includegraphics[trim={2.5cm .1cm 4.1cm 1cm},clip,width =.65\linewidth]{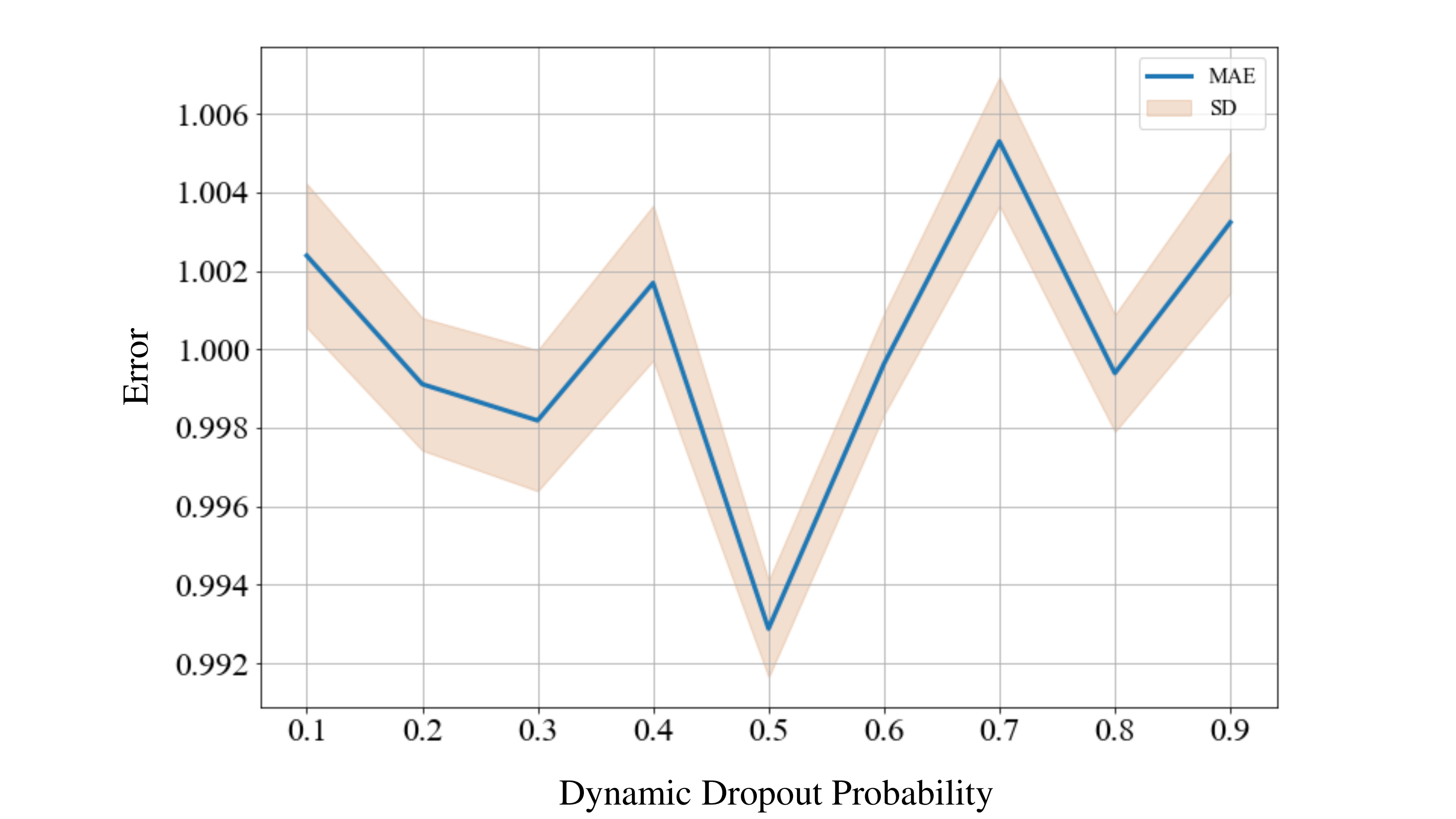}
 \caption{Effects of dynamic uncertainty optimization on prediction error and uncertainty during the occurrence of an anomaly. The method automatically selects the optimal probability that yields the lowest uncertainty.
}
 \label{fig_mcdrop}
\end{figure}

As an example, Figure~\ref{fig_mcdrop} shows that the optimal uncertainty results can occur when the standard deviation is the lowest. Intuitively, the network at $p=0.5$ shows the highest confidence in its prediction (i.e., the lowest uncertainty) where unnecessary neurons are dropped out from the network. Therefore, the network automatically selects and reports the $p=0.5$ probability as the best choice for this time step in the testing phase. 

\section{Experiments}
\label{section-experiment}

This section reports multiple experiments comparing the proposed AA-Forecast framework with baseline models using different types of large-scale time series datasets.

\subsection{Dataset and Experimental Settings}

Three real-world time series with diverse structures and domains are gathered\footnote{All datasets are publicly available at https://github.com/ashfarhangi/AA-Forecast}. The detailed description and data collection procedure are as follows:
\begin{itemize}
 \item[$\bullet$] We provide a new spatio-temporal benchmark dataset (\texttt{Hurricane}), which is suited for forecasting during extreme events and anomalies. The dataset is gathered through the Florida Department of Revenue which provides the monthly sales revenue (2003-2020) for the tourism industry in the 67 counties of Florida which are prone to annual hurricanes. We further enriched and aligned the raw time series with the history of hurricane categories for each region upon impact. More precisely, the hurricane category indicates the maximum sustained wind speed which can result in catastrophic damages~\cite{noaa}. 
 \item[$\bullet$] The second dataset (\texttt{COVID-19}) showcases the changes in the number of employees based on one million employees active in the US during the COVID-19 pandemic and is gathered from Homebase~\cite{homebaseData}. We further enriched the data with the state-level policies as an indication of extreme events (e.g., the state's business closure order).
 \item[$\bullet$] The third dataset (\texttt{Electricity}) is a publicly available benchmark dataset that contains the electricity consumption of 370 consumers on an hourly basis from 2011 to 2014. Note that this benchmark dataset is anonymized and does not contain extreme event labels, yet AA-Forecast is able to automatically extract the anomalies, indicating abrupt changes in trend and seasonality.
\end{itemize}

\begin{table}[!ht]
 \centering
 \caption{Summary statistics of the datasets.}
\normalsize	
\begin{tabular}{l|rrrrr} 
\hline
Dataset & Hurricane & COVID-19 & Electricity \\
\hline 
Time step & Monthly & Daily & Hourly \\
\# Unique time series & 9,876 & 15,312 & 370 \\
\# Observation & 9,876 & 15,312 & 11,952,480 \\
\# Train & 7,900 & 12,250 & 9,561,984 \\
\# Test & 1,975 & 3,062 & 2,390,496 \\
\# Regions & 48 & 50 & 370 \\
\# Extreme events & 88 & 100 & - \\
\# Anomalous points & 102 & 124 & 672 \\
\hline
\end{tabular}
 \label{tab:dataset}
\end{table}

We propose two sets of experiments for all baseline models. The first experiment follows a standard 80-20 dividing of the dataset to 
training and testing sets and $\tau=12$ for window length. 
The second experiment evaluates the zero-shot prediction capability of the model based on various window search ranges in \{3, 6, 12, 24\}, and thus is more applicable for real-world settings when the newly added time series cannot train on a newly added time series. Hence, the second experiment evaluates the prediction accuracy of all models on a set of completely unseen time series. 

The models are implemented using Python 3.7 and tested on a cloud workstation with two Intel Xeon 2.3 GHz CPUs, 64 GB RAM, and one Nvidia Tesla A100 GPU. We conduct a grid search over all tunable hyperparameters on the held-out validation set for baseline methods and our framework. 
To provide a fair evaluation, all baseline models benefit from the essential features extracted by AA-Forecast except the ARIMA model which does not benefit from multidimensional features. Moreover, future known information is not included in all the models.

We kept training to 40 iterations for all experiments. The reported values are the average of the observed error five times during the test stage. The hyperparameters of all baseline methods are tuned based on a grid search.

\begin{table}[!ht]
\centering
\caption{Hyperparameters of AA-Forecast used for each dataset.}{
\normalsize	
\begin{tabular}{lrrr}
\hline Parameter & Hurricane & COVID-19 & Electricity \\
\hline Batch size & 128 & 64 & 64 \\
Learning rate & $1 \times 10^{-5}$ & $3 \times 10^{-5}$ & $5 \times 10^{-5}$ \\
Weight decay & $1 \times 10^{-6}$ & $1 \times 10^{-5}$ & $1 \times 10^{-4}$ \\
Number of epochs & 40 & 40 & 40 \\
Static dropout & 0.5 & 0.4 & 0.6\\
\hline
\end{tabular} 
 \label{tab:hyper}}
 \vspace{-1mm}
\end{table}

\subsection{Methods for Comparison}
The baseline methods for comparison include:
\begin{itemize}
 \item[$\bullet$] ARIMA~\cite{ArimaOG}: A traditional autoregressive integrated moving average method for time series prediction and often used as a baseline.
 \vspace{1mm}
 \item[$\bullet$] AE-LSTM~\cite{aelstm}: An LSTM network that uses an autoencoder for deep feature extraction and provides a deterministic prediction. 
 \vspace{1mm}
 \item[$\bullet$] SARIMAX~\cite{SARIMAX}: An autoregressive model that can handle seasonality and exogenous features of time series. 
 \vspace{1mm}
 \item[$\bullet$] UberNN~\cite{uber}: An LSTM-based model that uses Monte Carlo dropout to provide uncertainty and is able to extract deep features of time series through autoencoders.
 \vspace{1mm}
 \item[$\bullet$] TSE-SC~\cite{cai2020traffic}: was recently proposed as a Transformer-based Deep Learning model that can forecast abrupt changes accurately. (i) STAR Decomposition to automatically ex-
tract essential features such as anomalies, (ii) an Anomaly-Aware Model to leverage such
extracted features, and (iii) a Dynamic Uncertainty Optimization to reduce the uncertainty of the network. The final predicted 
 \vspace{1mm}
 \item[$\bullet$] AA-Forecast (LSTM) is our proposed model with LSTM cells.
 \vspace{1mm}
 \item[$\bullet$] AA-Forecast (GRU) is our proposed model with GRU cells.
\vspace{-2mm}
\end{itemize}


\subsection{Metrics}
\vspace{-2mm}
For providing a comprehensive evaluation, we adopted three different evaluation metrics. 
The first evaluation metric is the Continuous Ranked Probability Score (CRPS), which evaluates probabilistic forecasting. Formally defined as
$\mathbf{C R P S}=\int_{-\infty}^{\infty}(F(y)-\mathds{1}(y-\hat{y}))^{2} \mathrm{~d}y$ where $F$ is the cumulative distribution function of its forecast distribution and $\mathds{1}$ is the Heaviside step function.
We also report the root mean square error (RMSE). Formally defined as $\mathbf{R M S E}=\sqrt{\frac{1}{N} \sum_{i=1}^{N}\left(y_{t,(i)}-\hat{y}_{t,(i)}\right)^{2}}$ where $y_{t}$ is the mean of the predicted distribution at time $t$ and $\hat{y}_{t}$ is the observed value at time $t$. The third evaluation metric is the standard deviation (SD) that is correlated to the uncertainty of the prediction and is denoted as $\mathbf{S D} = \sqrt{\frac{1}{N} \sum_{i=1}^{N}\left(y_{t,(i)}-\Tilde{y_t}\right)^{2}}$ where $\bar{y}$ is the mean of the predicted distribution.

\vspace{-2mm}
\subsection{Experimental Results}

We provide two comprehensive comparisons and evaluations of the proposed AA-Forecast framework: the aforementioned 80-20 testing where 20\% of the data are unseen, as well as the testing on zero-shot prediction where the whole time series is unseen.
In both cases, we calculate the CRPS, RMSE, and SD. Lastly, we provided an ablation study to discuss the effectiveness of different AA-Forecast components.

\noindent\textbf{The $\bm{80-20}$ testing.} We first used the `older' 80\% of each time series in training and tested the accuracy of prediction on the rest of 20\%.
Table~\ref{tab:table8020} reports the loss of the networks under such $80-20$ testing, where 
the SD of AA-Forecast (GRU) method is lower than all baseline methods, showing the model's high confidence in the forecasts.

\begin{table}[!ht]
\centering
\caption{Performance comparison of our proposed framework and baseline models under $80-20$ testing.}
\normalsize	
\begin{tabular}{ll|ccc}
& & \multicolumn{3}{c} {Dataset} \\
\hline Methods & Metrics & Electricity& COVID-19 & Hurricane
\\
\hline ARIMA~\cite{ArimaOG} & CRPS & 1.150 & 0.103 & 0.761 \\
& RMSE & 1.520 & 0.114 & 0.802 \\
& SD & 0.225 & 0.011 & 0.106 \\
\hline AE-LSTM~\cite{aelstm} & CRPS & 0.895 & 0.086 & 0.531\\
 & RMSE & 1.296 & 0.087 & 0.576 \\
& SD & 0.215 & 0.009 & 0.102 \\
\hline
SARIMAX~\cite{SARIMAX} & CRPS & 0.911 & 0.098 & 0.532 \\
& RMSE& 1.285 & 0.108 & 0.578 \\
 & SD & 0.195 & 0.009 & 0.093 \\
 \hline
UberNN~\cite{uber} & CRPS & 0.633 & 0.071 & 0.442\\
& RMSE & 1.015 & 0.081 & 0.453 \\
 & SD & 0.134 & 0.007 & 0.073 \\
 \hline
 TSE-SC~\cite{cai2020traffic} & CRPS & 0.583 & 0.062 & 0.384 \\
& RMSE & 0.983 & 0.072 & 0.423\\
& SD & 0.146 & 0.007 & 0.092\\
\hline \textbf{AA-Forecast} & CRPS & 0.546 & \textbf{0.059} & 0.237 \\
 \textbf{(LSTM)}& RMSE & 0.949 & \textbf{0.068} & 0.274 \\
& SD & 0.095 & \textbf{0.003} & 0.060 \\

\hline \textbf{AA-Forecast} & CRPS & \textbf{0.493} & 0.063 & \textbf{0.216} \\ 
 \textbf{(GRU)} & RMSE & \textbf{0.894} & 0.073 & \textbf{0.253} \\
 & SD & \textbf{0.081} & 0.003 & \textbf{0.051} \\ \hline
\end{tabular}
\label{tab:table8020}
\end{table}

Among the baseline methods, UberNN and TSE-SC have shown good accuracy but suffer from higher SD (uncertainty) compared to the AA-Forecast (LSTM-GRU) models. Considering that the extracted features are available for all the baseline methods, we believe the higher uncertainty of SD is due to their static dropout probability that is constant for all time steps. Therefore, the two proposed models, AA-Forecast (LSTM-GRU), consistently outperform state-of-the-art methods. Considering all three evaluation metrics, AA-Forecast (GRU) is the best-suited framework for our dataset as it provides higher accuracy and confidence.

\begin{figure}[!h]%
 \centering
 {{\includegraphics[trim={.1cm 0cm .18cm .1cm},clip,width =.7\linewidth]{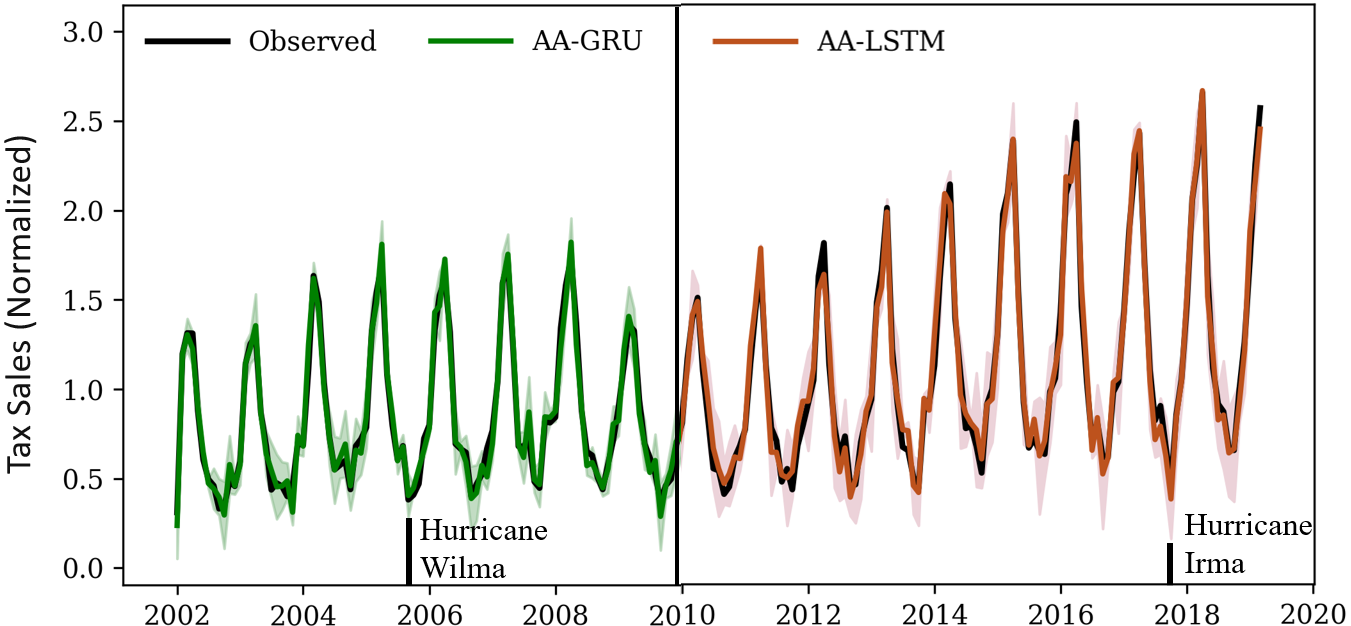} }}%
 \caption{Zero-shot prediction for hotel tax sales of Collier County, Florida, U.S. Both variations of AA-Forecast are concatenated for demonstration.
 }%
 \label{fig_unseenpred}%
\end{figure}

\noindent\textbf{Zero-shot prediction.} Table~\ref{tab:tableunseen} demonstrates the zero-shot prediction abilities for the selected models. Both AA-Forecast (LSTM-GRU) predictions follow the observed time series in general. The prediction errors are comparably low during the presence of extreme events (i.e., hurricanes). This is mainly due to the anomaly attention mechanism developed to further reduce the prediction error during extreme events. Moreover, extracted anomalies from STAR decomposition led to the recall of the hurricane effects on previously seen regions, thus providing predictions for unseen time series data with a lower error given the presence of anomalies. Figure~\ref{fig_unseenpred} showcases a sample of these predictions for each model where for every time step, the prediction uncertainty is the least.

\begin{table}[!h]
\centering
\caption{
Performance comparisons of zero-shot prediction abilities of models using ten randomly selected counties' sales tax data where they have not been used in training entirely.}
\centering
\begin{tabular}{ll|cccc}
& & \multicolumn{4}{c} {Input time window } \\
\hline Methods & Metrics & 3 & 6 & 12 & 24 \\
\hline ARIMA~\cite{ArimaOG} & CRPS & 0.893 & 0.891 & 0.861 & 0.831 \\ 
& RMSE & 0.934 & 0.932 & 0.922 & 0.872 \\ 
 & SD & 0.119 & 0.1154 & 0.115 & 0.113 \\ 
\hline AE-LSTM~\cite{aelstm} & CRPS & 0.663 & 0.661 & 0.651 & 0.601 \\ 
& RMSE & 0.708 & 0.706 & 0.696 & 0.646 \\ 
 & SD & 0.115 & 0.112 & 0.111 & 0.109 \\ 
\hline SARIMAX~\cite{SARIMAX} & CRPS & 0.664 & 0.662 & 0.662 & 0.602 \\ 
& RMSE & 0.714 & 0.712 & 0.712 & 0.652 \\ 
 & SD & 0.106 & 0.102 & 0.102 & 0.100 \\ 
\hline UberNN~\cite{uber} & CRPS & 0.547 & 0.545 & 0.535 & 0.485 \\ 
 & RMSE & 0.585 & 0.583 & 0.573 & 0.523 \\ 
 & SD & 0.086 & 0.082 & 0.082 & 0.08 \\ 
\hline TSE-SC~\cite{cai2020traffic} & CRPS & 0.766 & 0.764 & 0.754 & 0.704 \\ 
& RMSE & 0.795 & 0.793 & 0.783 & 0.733 \\ 
 & SD & 0.105 & 0.102 & 0.101 & 0.099 \\ 
\hline \textbf{AA-Forecast} & CRPS & 0.362 & 0.361 & 0.351 & 0.301 \\ 
\textbf{(LSTM)}& RMSE & 0.406 & 0.404 & 0.394 & 0.344 \\ 
 & SD & 0.073 & 0.071 & 0.069 & 0.067 \\ 
\hline \textbf{AA-Forecast } & CRPS & \textbf{0.348} & \textbf{0.346} & \textbf{0.336} & \textbf{0.286} \\ 
\textbf{(GRU)}& RMSE & \textbf{0.385} & \textbf{0.383} & \textbf{0.373} & \textbf{0.323} \\ 
 & SD & \textbf{0.064} & \textbf{0.060} & \textbf{0.062} & \textbf{0.058} \\ 
\hline
\end{tabular}

\label{tab:tableunseen}
\end{table}

Given that the network did not train on the selected time series directly, it's able to transfer its knowledge from previously seen extreme events (i.e., the effect of cat 4 hurricanes) and provide more accurate prediction when not provided with such ability.

\subsection{Ablation Study}
In this section, we provide an extensive analysis of the performance of AA-Forecast, as well as the impact of each component on the performance of AA-Forecast. The results are shown in Table~\ref{tab:table3} where we removed each component and reported the changes in accuracy and uncertainty.

\noindent\textbf{Influence of anomaly-aware decomposition.} To demonstrate that the anomaly-aware decomposition can aid in improving the time series prediction, we fed the input series to the prediction model directly. This modification results in the worst performance in our ablation study. Note that AA-Forecast (GRU) still benefits from dynamic dropout optimization and extreme event labels, and the predicted uncertainty is optimized. However, the accuracy of AA-Forecast prediction (GRU) drops because of the limited number of features, indicating that the neural network does not have a strong ability to capture complex and nonlinear information. This can highlight the role of auxiliary features such as decomposed anomalies and extreme events for forecasting.

\noindent\textbf{Influence of uncertainty optimization.}
We also used a static dropout throughout the experiments at every time step, which caused a substantial increase in SD. 
Uncertainty optimization of dropout plays a critical role in reducing the uncertainty of the forecast intervals. Such modification also caused a higher error in the forecast, which is the model's inability to forecast with higher confidence.

\noindent\textbf{Influence of anomaly attention.} We conducted experiments to demonstrate the effectiveness of anomaly-awareness through the network's attention mechanism. Specifically, we directly fed the extreme events and anomalies without the anomaly-attention mechanism described in Section~\ref{sec-Anomaly-Aware}. 
Such change makes limits AA-Forecast's knowledge about hurricanes and the severity of their effects. As an example, in Figure~\ref{fig_influence} (right), the results show that the network's error during the presence of harder-to-predict time points (anomalies and extreme events) weakens.
Thus, removing the attention mechanism for anomalous/extreme event points of the dataset will reduce the performance of the model during the critical months of extreme events such as hurricanes. Simply relying on the previously seen dataset will not allow the network to handle external events and sudden changes effectively. 

\begin{table}[!t]
 \caption{Ablation study on AA-Forecast (GRU) model using the sales tax dataset to show the effectiveness of its components.}
 \centering
\begin{tabular}{cc|cccc}

& & \multicolumn{4}{c} { Time window } \\
\hline AA-Forecast (GRU) & Metrics & 3 & 6 & 12 & 24 \\
\hline w/o STAR Decomposition & CRPS & 0.493 & 0.446 & 0.445 & 0.457 \\ 
& RMSE & 0.512 & 0.464 & 0.463 & 0.494 \\
& SD & 0.074 & 0.071 & 0.070 & 0.070 \\
\hline w/o Uncertainty Optimization & CRPS & 0.429 & 0.431 & 0.43 & 0.367 \\
& RMSE & 0.466 & 0.471 & 0.467 & 0.404 \\
& SD & 0.088 & 0.088 & 0.087 & 0.083 \\ 
\hline w/o Anomaly Attention & CRPS & 0.379 & 0.380 & 0.367 & 0.317 \\ 
& RMSE & 0.416 & 0.417 & 0.404 & 0.354 \\
 & SD & 0.067 & 0.067 & 0.063 & 0.061 \\
\hline AA-Forecast (GRU) & CRPS &\textbf{0.348} & \textbf{0.346} & \textbf{0.336} & \textbf{0.286} \\
 & RMSE & \textbf{0.385} & \textbf{0.383} & \textbf{0.373} & \textbf{0.323} \\
 & SD &\textbf{0.064} & \textbf{0.060} & \textbf{0.060} & \textbf{0.058} \\
\hline
\end{tabular}
\label{tab:table3}
\end{table}

\subsection{Discussion}
\begin{figure}[!t]%
 \centering
 {{\includegraphics[trim={3.4cm 1.5cm 4cm 1cm},clip,width =.6\linewidth]{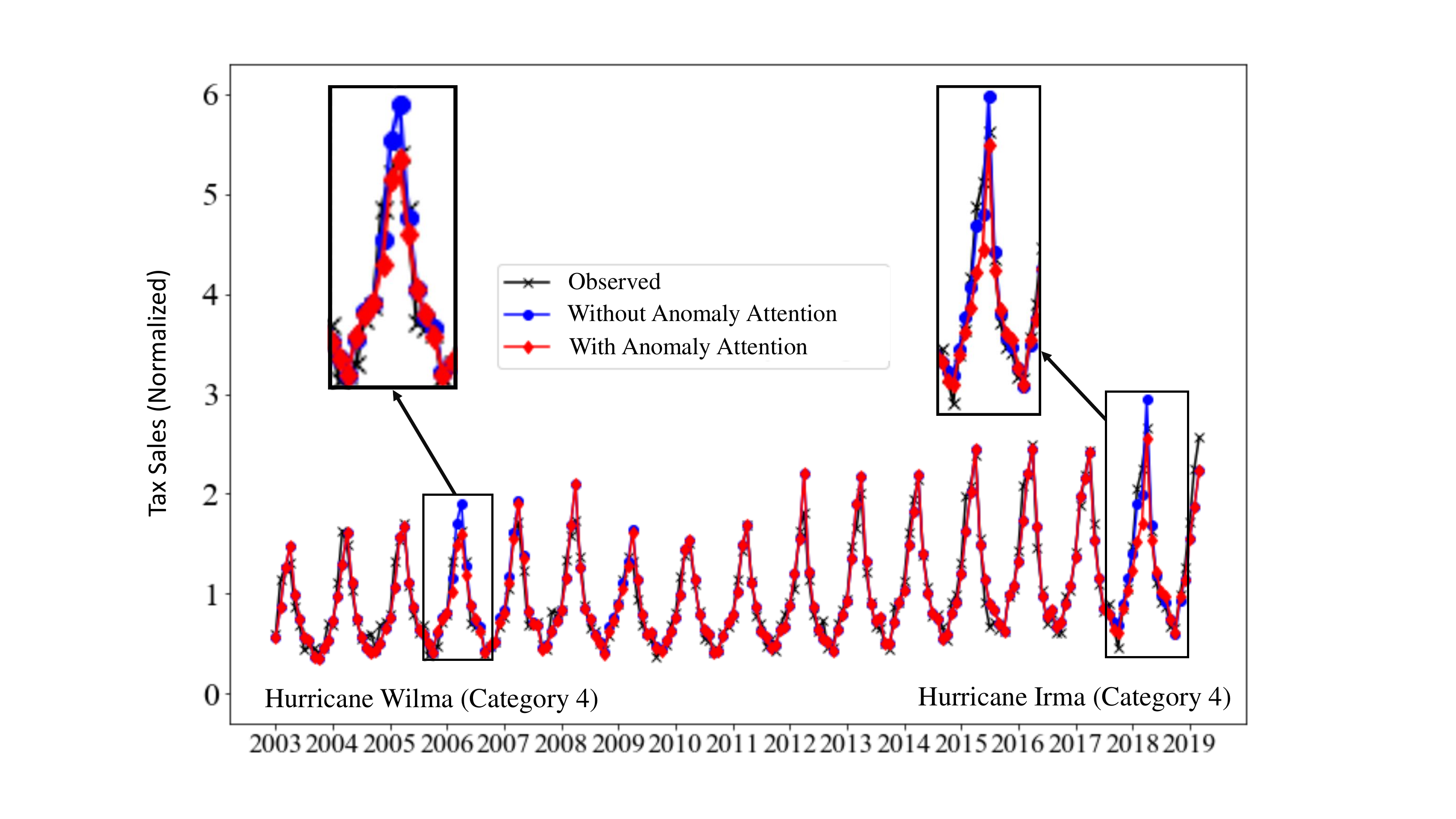} }}%
 \caption{Influence of anomaly attention on hurricanes. Two Category 4 hurricanes (Wilma and Irma) have caused similar annual sales losses. Anomaly-attention activation occurs during the presence of extreme events which makes it computationally efficient compared to the full-attention mechanism in transformers. }%
 \label{fig_influence}%
\end{figure}

\textbf{Interpretation.}
The benefits of providing optimal uncertainty in prediction are twofold: first, it provides a systematic way to aid in resource allocation. Second, it further prepares the domain for interventions. For example, if one region receives more catastrophic extreme events, the resources can be transferred to that region. Moreover, government and industries can provide better-informed interventions and decisions (e.g., financial aid relief during COVID-19).
As shown in the ablation study, including additional features such as extreme events and anomalous points can improve accuracy and better prime the model to handle predictions than deviate from trend or seasonality. Moreover, as shown in Figure~\ref{fig_influence} without proper attention to these points, they result in a large amount of error in forecasting. Given that such critical moments are of high importance during extreme events such as natural disasters, the performance of the model during critical time steps can be improved. Hence, it is essential to provide a thorough learning objective in our time series models to not only improve the overall performance but take critical moments into more consideration. Furthermore, allowing the model to provide its level of uncertainty establishes transparency and builds a level of trust for the users.

\noindent\textbf{Limitations \& future directions.} 
Although the dynamic dropout mechanism guarantees the least uncertainty in predictions, it cannot provide guarantees to do the same for prediction accuracy. This is due to the randomness nature of the dropout which we left as a future work where the dropout can appear for a predetermined distribution of the neurons. 
Therefore, maximizing the useful information contained in the multidimensional model serves to predict time series in extreme events. When it is not available, it's more reasonable to suggest methods that extract potential critical time steps such as anomalous points (e.g., STAR decomposition).

\section{Related Works}
\label{section-related}
Anomalies in time series data often produce a high variance of uncertainty prediction that is difficult to predict, thus becoming a challenge for reliable model design~\cite{uber,pang2017anomaly}. To provide a more reliable forecast during the presence of anomalies, probabilistic forecasting methods are often studied, which can report a level of uncertainty~\cite{IJCAI2019probalisiticModel}. 

The majority of Bayesian Neural Networks in probabilistic forecasting requires specific training and optimization methods and require additional model parameters that result in a larger amount of computation. Hence, MC dropout is preferred due to its practicability and its out-of-the-box solution~\cite{uber}. Applying standard dropout to Bayesian Neural Networks often results in poor performance on account of dropout noise preventing the network from maintaining long-term memory~\cite{mcsurvey}. 
Gal and Ghahramani~\cite{mcdropOG} proposed the MC dropout, in which the dropout can be interpreted as a sampling method that is equivalent to a variational approximation of a deep Gaussian process. MC dropout that is used for recurrent layers has proved to be successful and is commonly used in practice by applying dropout to recurrent connections in a way that can preserve long-term memory~\cite{mcsurvey}.
In previous studies, static MC dropout was used throughout their experiments, which suffers the model's robustness toward the effect of anomalies. Given that probabilistic models still require an overall great accuracy of their forecasts, optimizing the uncertainty in prediction intervals remains a challenging question

\section{Conclusion}
\label{section-conclusion}
We propose an anomaly-aware time series prediction framework, namely AA-Forecast, to capture and leverage the effect of extreme events and anomalies for the time series prediction task. It features a novel anomaly decomposition method that also extracts the essential features of the data. We also proposed an anomaly-aware model to leverage the extracted anomalies through an attention mechanism. Moreover, we reduced the uncertainty of the network without any further training so that the prediction uncertainty is minimal through the testing state. We compare our framework with several statistical and deep learning models using three real-world time series datasets. The results show that the AA-Forecast framework outperforms these models in prediction error and uncertainty. 
For future work, the prediction performance could be further improved if we target specific groups of neurons (e.g., the neurons containing unnecessary details of the time series dynamics) for dynamic dropout optimization.

\bibliographystyle{unsrt}  

\bibliography{Bibliography.bib}

\end{document}